\documentclass{article}

\usepackage{microtype}
\usepackage{graphicx}
\usepackage{booktabs}
\usepackage{hyperref}

\usepackage[preprint]{icml2026}

\usepackage{amsmath}
\usepackage{amssymb}
\usepackage{algorithm}
\usepackage{algorithmic}
\usepackage{listings}
\usepackage{xcolor}

\icmltitlerunning{How Should Agents Read Demonstrations?}

\begin{document}

\twocolumn[
  \icmltitle{How Should Agents Read Demonstrations?\\Hierarchical Structure Beats Flat Action Logs}

  \icmlsetsymbol{equal}{*}

  \begin{icmlauthorlist}
  \icmlauthor{Honjar Xing}{equal,mit}
  \icmlauthor{Jefferson Lin}{equal,mit}
  \icmlauthor{Henry Lieberman}{mit}
  \end{icmlauthorlist}

  \icmlaffiliation{mit}{MIT Computer Science and Artificial Intelligence Laboratory (CSAIL), Cambridge, MA, USA}

  \icmlcorrespondingauthor{Honjar Xing}{honjar@mit.edu}

  \icmlkeywords{programming by demonstration, LLM agents, hierarchical representations, human-centered agents, web automation}

  \vskip 0.3in
]

\printAffiliationsAndNotice{\icmlEqualContribution \\ Accepted at the 5th Deep Learning for Code (DL4C) Workshop, ICML 2026, Seoul, South Korea.}

% ============================================================
% ABSTRACT
% ============================================================
\begin{abstract}
Programming by Demonstration (PbD) offers a human-centered way to author procedural knowledge for LLM agents: users communicate what they want by showing rather than by writing prompts or code, making agent authoring accessible to non-programmers. The natural output of a PbD recording is a flat action log, but how this log is organized before being passed to the agent is an open design question with significant consequences for plan quality. We propose grouping recorded actions into labeled, hierarchical subgoals and evaluate the effect of this organizational structure in a controlled experiment. Across 85 web automation tasks, we compare a zero-shot baseline against four demonstration formats that share identical action sequences but differ in structure. On 43 natural-language tasks with vague descriptions, hierarchically grouped demonstrations improve pass rates from 76.7\% to 90.7\% (paired permutation test $p{=}0.034$; win-loss 6:0), while flat demonstrations show a smaller, non-significant improvement. On 42 tasks with precise descriptions, no format provides any benefit, confirming that the hierarchical advantage arises specifically when descriptions leave procedural details ambiguous. Ablation shows that subgoal grouping alone drives the effect: preconditions, postconditions, and parameter annotations add no measurable benefit. These results offer a concrete design recommendation for PbD pipelines and, more broadly, for any system that feeds procedural context to an LLM agent: segment action sequences into named subgoal groups rather than presenting flat step lists.
\end{abstract}

% ============================================================
% 1. INTRODUCTION
% ============================================================
\section{Introduction}
\label{sec:intro}

When an LLM agent faces a task it cannot solve from a natural-language description alone, it needs procedural guidance: step-by-step knowledge of how to accomplish the goal. The human-centered way to provide this guidance is Programming by Demonstration (PbD) \citep{cypher1993, lieberman2001}: a user performs the task while the system records their interactions, and the recording becomes a reusable procedure. PbD is appealing because it requires no programming knowledge and no ability to articulate precise specifications. Users communicate procedural intent by showing rather than by writing prompts or code. This is especially valuable for tasks with vague or preference-driven requirements where users know what they want but struggle to specify it in words.

Recent work has begun connecting PbD to LLM-based agents. ALLOY \citep{alloy} transforms user demonstrations into graph-structured workflows that LLM sub-task agents can execute and generalize, showing that demonstration-based authoring outperforms prompt-based agents in capturing user intent. AWM \citep{awm} induces reusable workflows from agent trajectories and retrieves them as in-context guidance for new tasks. AdaptAgent \citep{adaptagent} adapts multimodal web agents to new domains via a small number of human demonstrations. These systems each adopt a fixed demonstration representation with graph workflows, trajectory summaries, or raw action sequences, varying the content or quantity of demonstrations. However, none study how the organizational format of the same demonstration affects the agent's ability to follow it.

This is an important gap because there is strong evidence that format matters independently of content. \citet{he2024format} showed that presenting identical content as plain text, Markdown, JSON, or YAML shifts LLM accuracy by up to 40\%. \citet{min2022demos} demonstrated that the format of in-context demonstrations matters more than label correctness. \citet{skillsbench} found that curated, modular procedural documents improve agent performance by 16.2 percentage points while verbose, unstructured ones actively hurt. If format affects how well an LLM uses its context, then it should affect how well an agent follows a demonstrated procedure.

We test this hypothesis directly. We propose organizing PbD recordings into hierarchical subgoal groups. Rather than a flat numbered list of raw events, the post-processed demonstration groups actions under labeled subgoals (e.g., ``Navigate to slide,'' ``Open text box tool,'' ``Enter content'') each tagged with a functional type. This format, which we call the Agentic Goal Stack (AGS), mirrors the call-stack structure of procedural programs: each subgoal is a named operation with a defined scope. A deterministic post-processor converts the flat recording into this format automatically, requiring no additional user input.

We design a controlled experiment with five conditions that receive identical action sequences differing only in organizational format. We evaluate on 85 web automation tasks with 3 repetitions per condition (1,275 total executions). The agent generates executable code in the form of structured API call sequences via tool use, and a deterministic evaluator verifies end-state correctness. Our main findings are: (1)~hierarchically structured demonstrations improve pass rates on vague tasks from 76.7\% to 90.7\%, with every discordant task favoring hierarchy (win-loss 6:0); (2)~precisely specified tasks are unaffected, confirming the effect is specific to the setting where PbD is most needed; (3)~subgoal grouping alone accounts for the improvement. PbD pipelines with post-processors should segment recordings into labeled subgoal groups rather than emitting flat logs.

% ============================================================
% 2. RELATED WORK
% ============================================================
\section{Related Work}
\label{sec:related}

\paragraph{PbD for LLM agents.}
Several systems use demonstrations to improve agent performance, but none ablate the demonstration's organizational structure while holding the action sequence constant. ALLOY \citep{alloy} is the closest to our work: it transforms user demonstrations into graph-structured workflows that LLM sub-task agents execute, and a user study shows this outperforms prompt-based agents. However, ALLOY adopts a single representation and does not compare it against alternatives. AWM \citep{awm} induces reusable workflows from agent trajectories and retrieves them as context for new tasks. AdaptAgent \citep{adaptagent} adapts multimodal web agents to new domains via a small number of human demonstrations, either as in-context examples or by meta-learning a fine-tuning prior. LearnAct \citep{learnact} uses a multi-agent framework to extract knowledge from mobile-GUI demonstrations. Voyager \citep{voyager} builds a library of verified skill functions in executable code. Each of these systems fixes the demonstration representation while varying content or quantity. We hold the action sequence constant and vary only the organizational format.

\paragraph{Format sensitivity in LLMs.}
\citet{he2024format} showed that formatting identical content as plain text, Markdown, JSON, or YAML shifts LLM accuracy by up to 40\%. \citet{min2022demos} demonstrated that the format of in-context demonstrations matters more than label correctness for in-context learning. \citet{sclar2024} found that minor prompt variations can cause accuracy swings of over 70\% on some benchmarks. These results establish that how information is presented to an LLM matters independently of what information is presented, but prior format studies do not address agent demonstrations specifically.

\paragraph{Structured reasoning.}
A large body of work structures the agent's reasoning process rather than its input. Chain-of-Thought \citep{wei2022cot} adds intermediate reasoning steps; ReAct \citep{react} interleaves reasoning with actions; Tree of Thoughts \citep{tot} searches over alternative reasoning paths. These methods improve the generation process. Our work is complementary: we structure what the LLM \emph{sees} before generation, organizing recorded actions into typed subgoals.

\paragraph{PbD and web automation.}
Classic PbD systems \citep{cypher1993, lieberman2001} post-process recordings by inferring loops, generalizing values, and detecting repeated subsequences. Modern web-automation tools typically store recordings as flat action logs and replay them literally or pass them to an LLM. When recordings serve as context for plan generation rather than literal replay, the post-processing question returns: pass the raw log, or impose structure first? \citet{skillsbench} found that curated, modular procedural documents improve agent performance while verbose ones hurt, but did not propose a specific format. AGS provides one, designed to be generated automatically from PbD recordings. Benchmarks for web agents on productivity applications include WebArena \citep{webarena}, Mind2Web \citep{mind2web}, SheetCopilot \citep{sheetcopilot}, and PPTC \citep{pptc}. Our evaluation uses the Google Slides REST API for both execution and deterministic verification.

% ============================================================
% 3. HIERARCHICAL DEMONSTRATION FORMAT
% ============================================================
\section{Hierarchical Demonstration Format}
\label{sec:format}

We propose that an effective demonstration should encode structure beyond a flat action sequence. This section describes the four structural components, motivates the choice of a sequential hierarchy, and illustrates each experimental condition.

\subsection{Components}
\label{sec:components}

\paragraph{Subgoal hierarchy.}
A structured demonstration groups steps into named subgoals, each representing a coherent phase of the task. For instance, the seven raw actions for ``Bold the title text and change its font size to 36pt'' decompose into three subgoals: (1)~select the title element, (2)~apply bold formatting, and (3)~change the font size. Subgoal boundaries signal which steps form an atomic unit, preventing the agent from interleaving actions across phases.

\paragraph{Action typing.}
Each subgoal is tagged with one of three functional types: \textsc{process} (reusable UI manipulation: click, scroll, select-all), \textsc{input} (user-specific data entry: type text, set a value), or \textsc{output} (navigation or page-level state changes: open URL, switch tabs). This distinction tells the agent which steps to preserve literally and which values to adapt.

\paragraph{State annotations.}
Each subgoal carries a precondition (what must hold before execution) and a postcondition (what will hold after). For example, ``Apply bold'' has precondition ``Title text is selected'' and postcondition ``Title text is bold.'' These annotations let the agent verify expected state at subgoal boundaries. Flat step lists leave state expectations implicit.

\paragraph{Parameterization.}
Steps whose values are task-specific are marked with explicit parameter annotations. Rather than seeing \texttt{type("36")} as an opaque literal, the agent sees it annotated as \texttt{type(value=\{font\_size\}) \# PARAM}, with a parameter block listing the value and its type. This separates reusable scaffolding from task-specific data.

\subsection{Why Sequential Hierarchy?}
\label{sec:why_hierarchy}

The subgoal hierarchy is a sequential nesting with each subgoal containing an ordered list of actions and the subgoals themselves are ordered. This structure mirrors how UI procedures are naturally organized. A user navigates to a location, performs an operation at that location, and moves on. The resulting action sequence's call-stack structure includes low-level actions (keystrokes, clicks) are scoped within higher-level operations (``open menu search tool,'' ``format text''). Alternative structures such as directed graphs (allowing skip edges or parallel branches) or trees (allowing recursive subgoal decomposition) are more expressive but would require richer recordings or user annotation to populate. The sequential hierarchy captures the dominant structure in recorded demonstrations.

The post-processor segments a flat action sequence into subgoals using navigation boundaries (clicks on new UI regions, page transitions), action-type transitions (e.g., a navigation action followed by a data-entry action signals a phase boundary), and functional clustering (consecutive keystrokes that form a compound operation, such as a menu-search sequence, are grouped together). Each resulting subgoal is assigned a type (\textsc{process}, \textsc{input}, or \textsc{output}) based on its constituent actions. The segmentation is deterministic and requires no user input beyond the raw recording.

\subsection{Format by Condition}
\label{sec:conditions}

We evaluate five conditions on the same underlying action sequence. We illustrate with the task ``Bold the title text and change its font size to 36pt'' (7 actions).

\paragraph{C1 (Zero-shot).} No demonstration. The agent receives only the task description and must generate a plan from its own knowledge.

\paragraph{C2 (Flat demonstration).} A numbered list:

\begin{lstlisting}
DEMONSTRATION (7 steps):
1. CLICK on title text box
2. PRESS_KEY Ctrl+A
3. PRESS_KEY Ctrl+B
4. CLICK on font size dropdown
5. PRESS_KEY Ctrl+A
6. TYPE = "36"
7. PRESS_KEY Enter
\end{lstlisting}

\paragraph{C3 (Structured / AGS).} The full hierarchical format:

\begin{lstlisting}
AGS: bold-title-set-fontsize
GOAL: Bold title and set font size to 36pt

SUBGOAL 1: Select title element [PROCESS]
  Pre: Presentation open on slide 1
  Post: Title text fully selected
  1. click(target="title text box")
  2. press_key(key="ctrl+a")

SUBGOAL 2: Apply bold formatting [PROCESS]
  Pre: Title text fully selected
  Post: Title text is bold
  3. press_key(key="ctrl+b")

SUBGOAL 3: Set font size [INPUT]
  Pre: Title text still selected
  Post: Font size is 36pt
  4. click(target="font size dropdown")
  5. press_key(key="ctrl+a")
  6. type(value={font_size})  # PARAM
  7. press_key(key="Enter")

PARAMETERS:
  - font_size: "36" (type: number)
\end{lstlisting}

\paragraph{C4 (Hierarchy only).} Subgoal grouping and type tags, but no pre/postconditions or parameters. Tests whether subgoal boundaries alone account for the benefit.

\paragraph{C5 (Flat + type labels).} A flat list with each step prefixed by a type tag (\texttt{[INTERACTION]}, \texttt{[SHORTCUT]}, \texttt{[INPUT]}), but no subgoal grouping. Tests whether type labels help independently of hierarchy.

\subsection{Illustrative Example}
\label{sec:example}

We illustrate the format effect on NL17: ``Add contact information to slide~3.'' This task requires inserting a text box on a slide using the \texttt{TITLE\_ONLY} layout, which has no body placeholder. The required procedure involves navigating to slide~3, opening the text box tool via menu search (a 4-keystroke sequence: \texttt{Alt+/}, type ``Text box'', \texttt{ArrowDown}, \texttt{Enter}), drawing a text box, and typing the content.

Zero-shot (C1), the agent fails as it attempts to write into a nonexistent body placeholder. Under the flat format (C2), the 4-keystroke menu-search pattern appears as four unrelated steps, and the agent sometimes reproduces them incorrectly. Under the hierarchical format (C3), these keystrokes are grouped under ``Open text box tool [\textsc{process}],'' signaling a coherent, reusable pattern. A separate ``Draw and label text box [\textsc{input}]'' subgoal marks the typed content as task-specific. Under C3, the agent consistently succeeds.

% ============================================================
% 4. EXPERIMENTS
% ============================================================
\section{Experimental Evaluation}
\label{sec:experiments}

\subsection{Evaluation Pipeline}

We evaluate through an end-to-end automated pipeline (Algorithm~\ref{alg:pipeline}) that executes the agent's output and measures task success. For each task, the pipeline resets the application to a known state, provides the agent with a task description and (for C2--C5) a formatted demonstration, collects the agent's generated action plan, executes that plan against the live application via the Google Slides REST API, and then reads back the resulting application state to determine whether the task's success criteria are met with the API-measured end state either satisfies the criteria or not. 

Each task has an associated demonstration that is an action sequence specifying the interactions required to achieve the success criterion. A deterministic formatter converts the demonstration into the condition-specific representation. C1 receives no demonstration. The LLM (Claude Sonnet~4) generates executable code via the tool-use API, producing structured operations with typed parameters that are executed against the live application. The evaluator then reads the application state through the same API and checks it against the task's success criteria.

\begin{algorithm}[t]
\caption{Evaluation Pipeline}
\label{alg:pipeline}
\begin{algorithmic}[1]
\REQUIRE Task description $\ell$, condition $c \in \{C_1, \ldots, C_5\}$
\REQUIRE Demonstration $d$
\ENSURE Pass/fail evaluation against success criteria
\STATE \textsc{ResetState}() \hfill $\triangleright$ Restore to known state
\IF{$c = C_1$}
  \STATE $\text{ctx} \gets \ell$  \hfill $\triangleright$ Zero-shot
\ELSIF{$c = C_2$}
  \STATE $\text{ctx} \gets \ell \oplus \textsc{Flatten}(d)$  \hfill $\triangleright$ Flat demo
\ELSIF{$c = C_3$}
  \STATE $\text{ctx} \gets \ell \oplus \textsc{Structure}(d)$  \hfill $\triangleright$ Structured demo
\ELSIF{$c = C_4$}
  \STATE $\text{ctx} \gets \ell \oplus \textsc{HierarchyOnly}(d)$ \hfill $\triangleright$ Hierarchy only
\ELSIF{$c = C_5$}
  \STATE $\text{ctx} \gets \ell \oplus \textsc{FlatTyped}(d)$ \hfill $\triangleright$ Flat + type labels
\ENDIF
\STATE $a \gets \text{LLM}(\text{ctx})$ \hfill $\triangleright$ Tool-use API, forced choice
\STATE \textsc{API.Commit}($a$) \hfill $\triangleright$ Execute plan via REST API
\STATE \textbf{return} \textsc{Evaluator.Verify}() \hfill $\triangleright$ Deterministic check
\end{algorithmic}
\end{algorithm}

\paragraph{Statistical methods.} Each task runs 3 times per condition. We apply majority vote (${\geq}2/3$ reps = pass). We report McNemar's exact test and Cohen's~$h$ on majority-vote outcomes. We additionally report a paired permutation test on all raw binary observations (100,000 iterations, fixed seed), which does not require majority-vote aggregation.

\subsection{Task Suite}

We evaluate on 85 Google Slides tasks divided into two complementary suites.

\paragraph{Natural-language tasks (43).} These use vague, open-ended descriptions (e.g., ``make this slide look more professional,'' ``add a Q\&A slide at the end''). Multiple valid interpretations exist. This is precisely the setting where PbD is most valuable as an authoring method: users can show what ``professional'' means to them even when they cannot specify it in a prompt. The demonstration provides one concrete procedure, and the agent must parse and reproduce it. Evaluators verify the presence of reasonable changes rather than exact values: for example, ``make title and subtitle consistent'' checks that both share the same font size and weight, regardless of which values are chosen. This design isolates the format question: when the description is ambiguous and the demonstration carries the procedural intent, how well does the agent follow the demonstrated procedure under different organizational formats?

\paragraph{Structured tasks (42).} These use precise, deterministic descriptions with explicit success criteria. They span text editing (11), shapes (9), slides (7), composites (10), multi-step (2), and other categories (3), at three difficulty levels: Easy (13), Medium (22), and Hard (7). These tasks serve as a control: the agent can generate a correct plan from the description alone, so demonstrations should provide no benefit.

\subsection{Natural-Language Results}

\begin{figure}[t]
  \centering
  \includegraphics[width=\columnwidth]{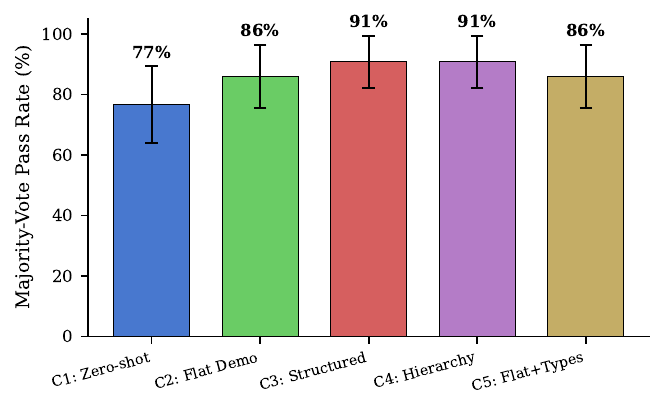}
  \caption{Pass rates on 43 natural-language tasks (3 reps, majority vote). Structured demonstrations (C3,~C4) improve pass rates from 76.7\% to 90.7\% (paired permutation test $p{=}0.034$; McNemar $p{=}0.031$; win-loss 6:0). Error bars show approximate 95\% confidence intervals.}
  \label{fig:nl}
\end{figure}

Figure~\ref{fig:nl} presents the primary result. On 43 natural-language tasks, flat demonstrations (C2) show a non-significant improvement over zero-shot (86.0\% vs.\ 76.7\% majority vote). Hierarchically structured demonstrations (C3) improve pass rates to 90.7\%, a 14.0pp improvement over zero-shot (McNemar $p{=}0.031$, Cohen's $h{=}0.39$). A paired permutation test on all 129 raw observations per condition (permuting condition labels within each task, 100,000 iterations, fixed seed) yields $p{=}0.034$ (two-sided), significant at $\alpha{=}0.05$. The win-loss record is 6:0: every task that changes outcome under structured demonstrations improves, and none degrades (Table~\ref{tab:nl_stats}). Hierarchy alone (C4) also achieves 90.7\%, matching the full AGS format.

\begin{table}[t]
\caption{Statistical tests: C1 vs.\ each condition on NL tasks (43 tasks). The primary comparison (C1 vs.\ C3) yields McNemar $p{=}0.031$ with a 6:0 win-loss record. Hierarchy alone (C4) matches full AGS.}
\label{tab:nl_stats}
\centering
\small
\begin{tabular}{@{}lcccc@{}}
\toprule
 & \textbf{C2} & \textbf{C3} & \textbf{C4} & \textbf{C5} \\
 & (Flat) & (AGS) & (Hier.) & (Types) \\
\midrule
Pass rate     & 86.0\% & 90.7\% & 90.7\% & 86.0\% \\
$\Delta$ from C1 & +9.3pp & +14.0pp & +14.0pp & +9.3pp \\
McNemar $p$   & 0.289  & 0.031  & 0.031  & 0.289  \\
Cohen's $h$   & 0.24   & 0.39   & 0.39   & 0.24   \\
Win--loss      & 6:2    & 6:0    & 6:0    & 6:2    \\
\bottomrule
\end{tabular}
\end{table}

The content of the demonstration (the procedural knowledge it provides) produces the first improvement: zero-shot to flat is $+$9.3pp. The format of the demonstration produces a further improvement: flat to hierarchical is $+$4.7pp. We discuss why this secondary improvement is practically significant despite its smaller magnitude in Section~\ref{sec:discussion}.

\paragraph{Why format matters on vague tasks.}
On vague tasks, the agent cannot infer the required procedure from the description alone. The demonstration supplies the missing procedural knowledge, but the flat format obscures which low-level actions form coherent operations. For instance, NL17 requires a 4-keystroke menu-search pattern that appears as four unrelated steps in the flat format. The hierarchical format groups these under a labeled subgoal, making the pattern parseable. NL30 (``add a Q\&A slide at the end'') scores 1/3 zero-shot but 3/3 under C3 because the \texttt{Ctrl+M} shortcut for creating a new slide, which the agent does not generate on its own, becomes visible under a labeled subgoal (``Create new slide [\textsc{output}]'').

\subsection{Structured Tasks: Control Condition}

\begin{figure}[t]
  \centering
  \includegraphics[width=\columnwidth]{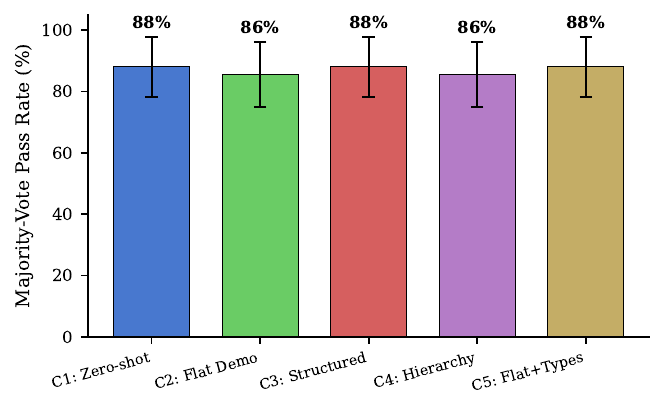}
  \caption{Pass rates on 42 structured tasks (3 reps, majority vote). All conditions cluster between 85.7\% and 88.1\%, with no significant differences. Demonstrations provide no benefit when the task description is already precise.}
  \label{fig:st}
\end{figure}

Figure~\ref{fig:st} shows that on 42 structured tasks, all five conditions cluster between 85.7\% and 88.1\% majority-vote pass rate. No condition differs significantly from zero-shot (all McNemar $p{=}1.00$; win-loss counts balanced at 3:3 or 3:4). When the task description specifies exactly what to do, the agent generates a correct plan without procedural guidance. This confirms that the NL improvement is not a general tendency to perform better with any demonstration context, and it establishes that demonstrations help specifically when descriptions leave the procedure ambiguous.

% ============================================================
% 5. DISCUSSION
% ============================================================
\section{Discussion}
\label{sec:discussion}

\subsection{Which Components Drive the Effect?}

AGS bundles four components: subgoal hierarchy, action typing, state annotations, and parameterization. Our two ablation conditions isolate their contributions.

\paragraph{Hierarchy is the active ingredient.}
C4 (hierarchy without annotations) matches C3 (full AGS) at 90.7\% on NL tasks and is comparable on structured tasks (85.7\% vs.\ 88.1\%). Stripping preconditions, postconditions, and parameter annotations from the hierarchical format has no measurable effect on plan quality.

\paragraph{Type labels alone do not help.}
C5 (flat list with action-type tags) scores identically to C2 (flat demo) at 86.0\%. Adding \texttt{[INTERACTION]}, \texttt{[SHORTCUT]}, or \texttt{[INPUT]} prefixes to a flat step list does not improve plan generation. The information provided by type labels is apparently only useful when it accompanies subgoal boundaries.

\paragraph{Hierarchy resolves flat-format ambiguity.}
On NL11 (``make title and subtitle consistent''), both flat formats (C2,~C5) fail while both hierarchical formats (C3,~C4) succeed. The flat demonstration presents steps where operations on the title and operations on the subtitle are interleaved at the same level, and the agent confuses them. Subgoal grouping separates these into distinct phases. One might observe that the content effect (zero-shot to flat: $+$9.3pp) is larger than the format effect (flat to hierarchical: $+$4.7pp) and conclude that format is secondary. Since the 6:0 win-loss record means hierarchy never hurts, hierarchy provides a safety margin against this failure mode.

\begin{table}[t]
\caption{Format ablation: win-loss counts (majority vote, discordant pairs only) across all 85 tasks.}
\label{tab:ablation}
\centering
\small
\begin{tabular}{@{}lcc@{}}
\toprule
Comparison & Win-Loss & Interpretation \\
\midrule
C4 vs.\ C2 & 3:1 & Hierarchy $>$ flat \\
C4 vs.\ C5 & 2:1 & Hierarchy $>$ type labels \\
C3 vs.\ C4 & 2:1 & Annotations add nothing \\
C5 vs.\ C2 & 1:0 & Type labels $\approx$ flat \\
\bottomrule
\end{tabular}
\end{table}

\subsection{Implications for Agent Context Design}

The finding that hierarchical subgoal grouping helps LLM agents parse procedural context extends beyond PbD recordings. Any system that feeds demonstrations, workflows, or procedural documentation to a code-generating agent faces the same design question: flat or structured? This includes coding agents that consume documentation or examples before generating executable plans. Our results suggest that systems which retrieve and present procedural context to agents should organize that context into labeled subgoal groups rather than flat sequences.

The concentration of the effect on vague natural-language tasks is especially relevant for human-centered agent design. These tasks represent the core use case for PbD: the user knows what they want but cannot articulate it precisely, so they demonstrate instead. Our results show that not only does PbD help in this setting (flat demonstrations improve over zero-shot), but that hierarchical post-processing makes PbD substantially more effective (structured demonstrations improve over flat). This suggests that the post-processing step is a critical, underexplored component of the PbD-to-agent pipeline.

\subsection{Limitations}
\label{sec:limitations}

All experiments use a single LLM (Claude Sonnet~4) on a single domain (Google Slides). Different models may respond differently to organizational structure. The Google Slides REST API is used for both execution and evaluation, which avoids browser-based noise but restricts the task space to API-expressible operations. Our demonstrations were recorded and verified through the PbD pipeline, which captures real user interactions and segments them into action sequences. Future work should evaluate with multiple LLMs and additional web applications.

% ============================================================
% 6. CONCLUSION
% ============================================================
\section{Conclusion}

Hierarchically structured demonstrations improve agent pass rates on vague tasks from 76.7\% to 90.7\% (McNemar $p{=}0.031$; paired permutation test $p{=}0.034$; win-loss 6:0), while flat demonstrations and precisely-specified controls show no significant effect. Subgoal grouping alone accounts for the improvement; annotations and parameter markup add nothing measurable. The design recommendation is simple: PbD post-processors should segment recorded actions into labeled subgoal groups rather than emitting flat logs. Even a lightweight segmentation heuristic such as grouping steps by navigation boundaries or temporal gaps may capture most of the benefit, since the ablation shows that the subgoal boundaries themselves, not the metadata attached to them, drive the effect.

\section*{Acknowledgments}
This work was supported by the US Air Force Office of Scientific Research under a grant entitled ``Architectures for Cognitive Intelligence.''
\bibliography{references}
\bibliographystyle{icml2026}

% ============================================================
% APPENDIX
% ============================================================
\newpage
\appendix

\section{Statistical Methodology}
\label{app:stats}

\paragraph{Experimental design.} We evaluate all 5 conditions on both task suites: 43 NL tasks and 42 structured tasks, for a total of 85 tasks $\times$ 5 conditions $\times$ 3 repetitions $=$ 1,275 executions.

\paragraph{Majority vote.} Each task runs 3 times per condition and passes if ${\geq}2/3$ repetitions succeed.

\paragraph{Win-loss.} A ``win'' is a task that fails C1 but passes the comparison condition; a ``loss'' is the reverse. Tasks passing or failing both are excluded.

\paragraph{Interpretation.} On NL tasks, McNemar's exact test yields $p{=}0.031$ for C1 vs.\ C3. All six discordant NL tasks favor C3 (win-loss 6:0). A paired permutation test on all 129 raw binary observations per condition (permuting condition labels within each task, 100,000 iterations, fixed random seed) yields $p{=}0.034$ (two-sided), significant at $\alpha{=}0.05$. On structured tasks, no condition significantly differs from C1 ($p{=}1.00$). The permutation test code and raw data are available as supplementary material.

\section{Per-Task Natural-Language Results}
\label{app:nl_results}

Table~\ref{tab:nl_pertask} provides complete per-task results for all 43 NL tasks.

\begin{table*}[t]
\caption{Per-task NL results (pass count out of 3 reps). Pattern classification uses majority vote.}
\label{tab:nl_pertask}
\centering
\scriptsize
\begin{tabular}{@{}llcccccl@{}}
\toprule
Task & Description & C1 & C2 & C3 & C4 & C5 & Pattern \\
\midrule
NL01 & Make title stand out & 3/3 & 3/3 & 3/3 & 3/3 & 3/3 & Both pass \\
NL02 & Closing thank-you slide & 3/3 & 3/3 & 3/3 & 3/3 & 3/3 & Both pass \\
NL03 & Red italic subtitle & 3/3 & 3/3 & 3/3 & 3/3 & 3/3 & Both pass \\
NL04 & Blue titles all slides & 2/3 & 3/3 & 3/3 & 3/3 & 3/3 & Both pass \\
NL05 & Text box with team name & 3/3 & 3/3 & 3/3 & 3/3 & 3/3 & Both pass \\
NL06 & Colorful accent shape & 3/3 & 3/3 & 3/3 & 3/3 & 3/3 & Both pass \\
NL07 & Four-slide deck & 3/3 & 3/3 & 3/3 & 3/3 & 3/3 & Both pass \\
NL08 & Bulleted body text & 3/3 & 3/3 & 3/3 & 3/3 & 3/3 & Both pass \\
NL09 & Center-align all text & 3/3 & 3/3 & 3/3 & 3/3 & 3/3 & Both pass \\
NL10 & Quarterly summary table & 3/3 & 3/3 & 3/3 & 3/3 & 3/3 & Both pass \\
NL11 & Consistent title+subtitle & 3/3 & 0/3 & 2/3 & 2/3 & 0/3 & Flat-fmt ambig. \\
NL12 & Visual separator & 3/3 & 3/3 & 3/3 & 3/3 & 3/3 & Both pass \\
NL13 & Split into topic slides & 3/3 & 3/3 & 3/3 & 3/3 & 3/3 & Both pass \\
NL14 & Look professional & 3/3 & 1/3 & 2/3 & 3/3 & 1/3 & Flat-fmt ambig. \\
NL15 & Speaker notes re: quarters & 3/3 & 3/3 & 2/3 & 3/3 & 3/3 & Both pass \\
NL16 & Highlight subtitle & 3/3 & 3/3 & 3/3 & 3/3 & 3/3 & Both pass \\
NL17 & Contact info on slide 3 & 1/3 & 3/3 & 3/3 & 3/3 & 3/3 & Hierarchy wins \\
NL18 & Duplicate + rename title & 0/3 & 3/3 & 3/3 & 3/3 & 3/3 & Hierarchy wins \\
NL21 & Underline subtitle & 3/3 & 3/3 & 3/3 & 3/3 & 2/3 & Both pass \\
NL24 & Bold all titles & 2/3 & 3/3 & 3/3 & 3/3 & 3/3 & Both pass \\
NL27 & Sans-serif font all slides & 0/3 & 3/3 & 3/3 & 2/3 & 2/3 & Hierarchy wins \\
NL28 & Different color subtitle & 3/3 & 3/3 & 3/3 & 3/3 & 3/3 & Both pass \\
NL30 & Q\&A slide at end & 1/3 & 3/3 & 3/3 & 3/3 & 3/3 & Hierarchy wins \\
NL31 & Strikethrough subtitle & 3/3 & 3/3 & 3/3 & 3/3 & 3/3 & Both pass \\
NL32 & Background color & 1/3 & 1/3 & 0/3 & 0/3 & 1/3 & Both fail \\
NL33 & Agenda slide & 3/3 & 3/3 & 3/3 & 3/3 & 3/3 & Both pass \\
NL34 & Easier to read body & 0/3 & 0/3 & 0/3 & 0/3 & 0/3 & Both fail \\
NL35 & Reduce title size & 3/3 & 3/3 & 3/3 & 3/3 & 3/3 & Both pass \\
NL36 & Numbered list on body & 1/3 & 3/3 & 2/3 & 3/3 & 3/3 & Hierarchy wins \\
NL37 & Move last slide to second & 3/3 & 3/3 & 3/3 & 3/3 & 3/3 & Both pass \\
NL38 & Remove middle slide & 2/3 & 2/3 & 3/3 & 3/3 & 3/3 & Both pass \\
NL39 & Decorative shape & 3/3 & 3/3 & 3/3 & 3/3 & 3/3 & Both pass \\
NL40 & Green subtitle text & 3/3 & 3/3 & 3/3 & 3/3 & 3/3 & Both pass \\
NL41 & Italic title text & 3/3 & 2/3 & 3/3 & 3/3 & 3/3 & Both pass \\
NL42 & Bulleted list slide 3 & 1/3 & 0/3 & 0/3 & 0/3 & 0/3 & Both fail \\
NL43 & Center-align subtitle & 3/3 & 3/3 & 3/3 & 3/3 & 3/3 & Both pass \\
NL44 & Blank slide at end & 1/3 & 3/3 & 3/3 & 3/3 & 3/3 & Hierarchy wins \\
NL45 & Bold subtitle text & 3/3 & 2/3 & 3/3 & 3/3 & 3/3 & Both pass \\
NL46 & Replace title with Welcome & 3/3 & 3/3 & 3/3 & 3/3 & 3/3 & Both pass \\
NL47 & Duplicate slide 2 & 3/3 & 3/3 & 3/3 & 3/3 & 3/3 & Both pass \\
NL48 & Left-align body text & 0/3 & 0/3 & 0/3 & 0/3 & 0/3 & Both fail \\
NL49 & Speaker notes slide 2 & 3/3 & 3/3 & 3/3 & 3/3 & 3/3 & Both pass \\
NL50 & Underline all titles & 3/3 & 3/3 & 3/3 & 3/3 & 3/3 & Both pass \\
\bottomrule
\end{tabular}
\end{table*}

\section{Structured Task Behavioral Patterns}
\label{app:patterns}

The 42 structured tasks fall into seven behavioral patterns (Table~\ref{tab:patterns}). Most (31/42) pass all conditions. Three are demo-enabled: the demonstration teaches UI paths the agent cannot discover zero-shot. Two exhibit flat-format ambiguity: they pass zero-shot but fail under flat demo, then recover under structured demo. Two contain non-generalizable content in the demonstration (coordinates or colors that do not match the target task). Two fail all conditions.

\begin{table}[h]
\caption{Behavioral patterns across 42 structured tasks (majority vote), with representative per-condition pass counts.}
\label{tab:patterns}
\centering
\scriptsize
\begin{tabular}{@{}lccccccc@{}}
\toprule
Pattern & N & C1 & C2 & C3 & C4 & C5 \\
\midrule
All pass         & 31 & 3/3 & 3/3 & 3/3 & 3/3 & 3/3 \\
Demo-enabled     & 3  & 1/3 & 3/3 & 3/3 & 3/3 & 3/3 \\
Flat-fmt ambig.  & 2  & 3/3 & 0\text{--}1/3 & 2\text{--}3/3 & 1\text{--}3/3 & 0\text{--}2/3 \\
Non-gen.\ content & 2  & 3/3 & 0/3 & 0/3 & 0/3 & 0/3 \\
All fail         & 2  & 0/3 & 0/3 & 0/3 & 0/3 & 0/3 \\
Other            & 2  & 3/3 & 3/3 & 1\text{--}3/3 & 1\text{--}3/3 & 2\text{--}3/3 \\
\bottomrule
\end{tabular}
\end{table}

\end{document}